\documentclass[runningheads]{llncs}
\usepackage[T1]{fontenc}
\usepackage{graphicx}
\usepackage[labelfont=bf,font=small,tableposition=bottom]{caption}
\usepackage[skip=3pt]{subcaption}
\usepackage{amsmath}

\usepackage{booktabs}
\usepackage{multirow}
\usepackage{tabularx}
\usepackage{array}
\usepackage{makecell}
\usepackage{xurl}
\usepackage{etoolbox}

\usepackage{acronym}

\acrodef{vo}[VO]{visual odometry}
\acrodef{slam}[SLAM]{simultaneous localization and mapping}
\acrodef{gpu}[GPU]{graphic processor unit}
\acrodef{mla}[MLA]{micro lens array}
\acrodef{sfm}[SfM]{structure from motion}
\acrodef{rmse}[RMSE]{root mean square error}
\acrodef{sd}[SD]{standard deviation}
\acrodef{dof}[DoF]{degrees of freedom}
\acrodef{nvs}[NVS]{novel view synthesis}

\begin{document}
\title{LiFMCR: Dataset and Benchmark for Light Field Multi-Camera Registration}
\titlerunning{LiFMCR}
%
\newcommand{\repthanks}[1]{\textsuperscript{\ref{#1}}}
\makeatletter
\patchcmd{\maketitle}
  {\def\thanks}
  {\let\repthanks\repthanksunskip\def\thanks}
  {}{}
\patchcmd{\@maketitle}
  {\def\thanks}
  {\let\repthanks\@gobble\def\thanks}
  {}{}
\newcommand\repthanksunskip[1]{\unskip{}}
\makeatother
\author{Aymeric Fleith\thanks{These authors contributed equally.\protect\label{X}}\inst{,1,2} \and
Julian Zirbel\repthanks{X}\inst{,1,2} \and 
Daniel Cremers\inst{1} \and
Niclas Zeller\inst{2}}

\authorrunning{Fleith and Zirbel et al.}
%
\institute{Technical University of Munich, Munich, Germany\\
\email{\{aymeric.fleith, julian.zirbel, cremers\}@tum.de}\\
\and Karlsruhe University of Applied Sciences, Karlsruhe, Germany\\
\email{niclas.zeller@h-ka.de}}

\maketitle              
\begin{abstract}
We present LiFMCR, a novel dataset for the registration of multiple \ac{mla}-based light field cameras. While existing light field datasets are limited to single-camera setups and typically lack external ground truth, LiFMCR provides synchronized image sequences from two high-resolution Raytrix R32 plenoptic cameras, together with high-precision 6-\ac{dof} poses recorded by a Vicon motion capture system. This unique combination enables rigorous evaluation of multi-camera light field registration methods.

As a baseline, we provide two complementary registration approaches: a robust 3D transformation estimation via a RANSAC-based method using cross-view point clouds, and a plenoptic PnP algorithm estimating extrinsic 6-\ac{dof} poses from single light field images. Both explicitly integrate the plenoptic camera model, enabling accurate and scalable multi-camera registration. Experiments show strong alignment with the ground truth, supporting reliable multi-view light field processing.

Project page: \url{https://lifmcr.github.io/}.


\end{abstract}

\section{Introduction}
\label{sec:introduction}

\begin{figure}[t]
  \centering
  \includegraphics[width=0.83\linewidth]{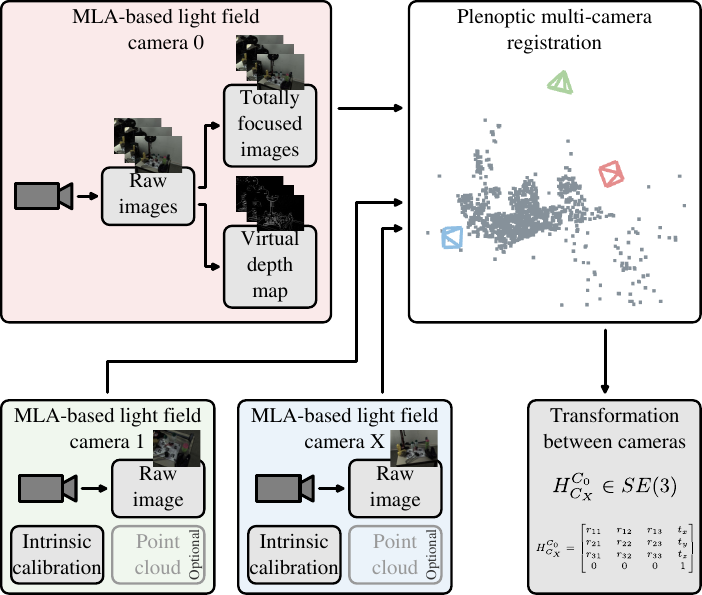}

   \caption{Pipeline for registering camera images and estimating 6-DOF extrinsics between views. Camera 0 serves as reference and other cameras are registered using one of the two proposed methods. Note that the point clouds from cameras 1 to X are required only for the RANSAC method, not for the plenoptic PnP algorithm.}
   \label{fig:pipeline}
\end{figure}

Accurate 3D reconstruction is essential for autonomous systems and robotic applications~\cite{engel2018direct,mur2017orb}. In contexts where precision, reliability, and adaptability are crucial, advanced imaging technologies such as \acf{mla}-based light field cameras, called plenoptic cameras in the sequel, offer advantages by capturing both spatial and angular information of light rays. Plenoptic cameras have been used for \acf{vo} and \acf{slam}~\cite{zeller2018spo}, depth estimation~\cite{wang2022occlusion}, super-resolution~\cite{xiao2023cutmib}, and post-capture refocusing~\cite{hahne2016refocusing}. While a single camera enables depth estimation from one image, combining multiple plenoptic cameras extends depth range and accuracy through stereo benefits. Reliable 3D reconstruction thus depends on robust plenoptic multi-camera calibration to align views within a consistent geometry.

In this paper, we introduce a new dataset with a benchmark of two methods for the 6~\ac{dof} registration of plenoptic multi-camera setups, explicitly addressing their optical and geometric challenges. To the best of our knowledge, no public datasets provide synchronized multi-view light field data together with external ground truth, limiting the evaluation of registration methods. Our dataset fills this gap, enabling rigorous benchmarking of plenoptic multi-camera registration. The included methods, which integrate the plenoptic camera model, ensure accurate spatial alignment across viewpoints --- an essential capability for enhancing depth perception, expanding the field-of-view, and improving robustness to occlusions in robotic applications.

Building on this foundation, our work enables more comprehensive benchmarking to advance reliable 3D perception with plenoptic cameras in applications such as autonomous navigation, human-robot interaction, and industrial inspection. The paper introduces the following key contributions:
\begin{itemize}
    \item A new plenoptic multi-camera dataset that provides synchronized sequences from two high resolution plenoptic cameras and sub-millimeter ground truth 6-\ac{dof} poses.
    \item A complete pipeline for intrinsic and extrinsic calibration of a plenoptic multi-camera setup.
    \item Two plenoptic camera registration benchmark algorithms to determine the relative 6-\ac{dof} poses of multiple cameras: a solution based on 3D transformation estimation via RANSAC, and the first algorithm to apply PnP to plenoptic data using a single image for registration.
\end{itemize}

The paper is organized as follows. Sec.~\ref{sec:relatedWork} presents related work on plenoptic camera and multi-camera calibration. Sec.~\ref{sec:method} introduces two benchmark registration methods to obtain the extrinsic calibration between plenoptic cameras. The contents of the proposed dataset and its acquisition method are explained in Sec.~\ref{sec:dataset}. Sec.~\ref{sec:evaluation} provides an extensive evaluation of both methods on the provided dataset. Finally, Sec.~\ref{sec:conclusion} summarizes and concludes the work.

\section{Related Work}
\label{sec:relatedWork}

This section reviews plenoptic camera calibration and the calibration of multi-camera systems. An overview of existing datasets for plenoptic cameras is also provided to highlight current limitations and the need for improved multi-view benchmarks.

\subsection{Plenoptic Cameras Calibration}

The two main configurations of light field cameras are unfocused plenoptic cameras (plenoptic camera 1.0) and focused plenoptic cameras (plenoptic camera 2.0). Each presents distinct challenges for modeling and calibration.
 
\textbf{Unfocused Plenoptic Camera Calibration:} In the unfocused plenoptic camera configuration~\cite{ng2005light}, the main lens is focused on the \ac{mla}, which itself is focused at infinity. The sensor plane is positioned at the focal plane of the \ac{mla}.

Assemblies of this type have been extensively studied in the literature. Calibration methods typically rely on processing reconstructed images, such as sub-aperture images, to enable reliable feature detection~\cite{dansereau2013decoding,zhou2019two,darwish2019plenoptic}. Alternatively, features can be detected directly in micro images~\cite{bok2017geometric,o2018calibrating,zhao2020metric}.

However, this type of plenoptic camera tends to be less commonly used because of a limited lateral resolution. Moreover, these calibration methods are generally not applicable to focused plenoptic cameras.

 \textbf{Focused Plenoptic Camera Calibration:} In the focused plenoptic camera configuration~\cite{lumsdaine2009focused,perwass2012single} the \ac{mla} is in front of or behind the image plane of the main lens. The micro lenses are focused at this image plane.

This configuration has led to new calibration methods, including a projection model with metric calibration~\cite{johannsen2013calibration,heinze2016automated} and several light-field-based models~\cite{zeller2016depthcalib,zeller2017plencalib,zeller2017dpo} for full intrinsic, extrinsic, and scene parameter estimation. Most approaches require image reconstruction, though some operate directly on raw images~\cite{labussiere2020blur,labussiere2022leveraging,zhang2016decoding,noury2017light}.

These methods often rely on calibration targets, which are cumbersome and require specialized equipment. To address this, the approach in~\cite{fachada2022pattern} extends the method of~\cite{fachada2021calibration} by treating sub-aperture views as pinhole views to avoid reference patterns, while the work in~\cite{fleith2024lifcal} enables recalibration on arbitrary scenes.

\subsection{Multi-Camera Calibration}

Several approaches have been proposed to determine the relative pose between cameras. Conventional methods rely on known calibration patterns, while more advanced techniques exploit the environment structure. Recent work also explores deep learning–based methods to estimate poses directly from images.

\textbf{Pattern-Based Calibration:} Multi-camera systems can be calibrated using known patterns observed by all cameras with overlapping fields of view~\cite{li2013multiple}. Non-overlapping cameras can be calibrated by adding a temporary third camera~\cite{robinson2017robust} or using a mirror to create overlap~\cite{kumar2008simple}. In this way, the calibration target method imposes a constraint to the cameras' field of view.

\textbf{Environment-Based Calibration:} Environment-based methods use scene features instead of calibration targets for greater flexibility. First, extrinsic calibration was performed by matching environmental points with a \ac{slam} reconstruction~\cite{carrera2011slam}, later simplified using a high-accuracy map and P3P~\cite{heng2015leveraging}. \cite{lin2020infrastructure} presents a similar approach without requiring intrinsic calibration. Moving elements of the scene are used as features~\cite{xu2021wide}. The advantage of these methods is that they can be used in situations where regular recalibration is required.

\textbf{Deep Learning-Based Calibration:} Deep learning methods estimate camera poses directly, first framed as end-to-end regression~\cite{kendall2015posenet}, later improved with new architectures~\cite{wu2017delving,naseer2017deep} to predict pose from a single view. Other approaches regress relative pose from image pairs~\cite{melekhov2017relative,laskar2017camera}.

\subsection{Datasets of Plenoptic Cameras}

Several plenoptic datasets have been released following the introduction of the Lytro and Lytro Illum cameras~\cite{rerabek2016new,paudyal_smart_2016}. However, most lack ground truth data and multi-camera setups.
More recent datasets primarily use unfocused plenoptic camera configurations but still lack proper synchronization and external ground truth~\cite{sancho_3dcuration_2024,palmieri_multi-focus_2018}. A later multi-camera dataset addressed this limitation by incorporating robot-based ground truth~\cite{sancho_bigmouth_2024}, yet the baseline configuration remained rigid. Sec.~\ref{sec:dataset} provides more details on existing datasets.

\section{Multi-Camera Registration}
\label{sec:method}

\begin{figure}[t]
  \centering
  \begin{subfigure}{0.47\linewidth}
    \includegraphics[width=\linewidth]{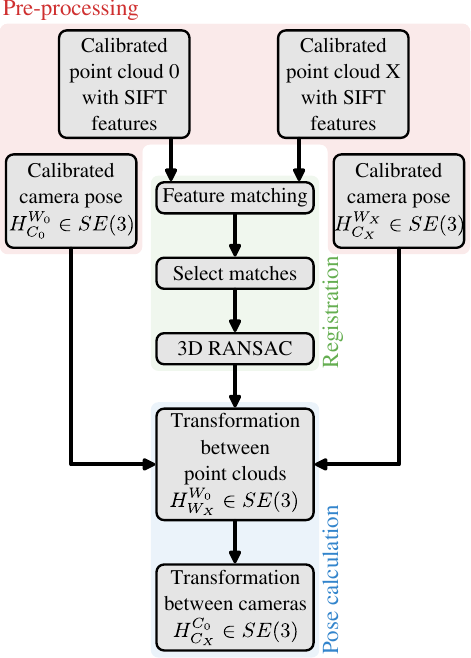}
    \caption{3D RANSAC plenoptic multi-camera registration. The reference camera 0 and the cameras to be registered (from 1 to X) must acquire a sequence of the scene to obtain an initial calibration and a point cloud. The registration is performed using 3D RANSAC before calculating the 6-\ac{dof} pose of each camera.}
    \label{fig:pipeline_RANSAC}
  \end{subfigure}
  \hfill
  \begin{subfigure}{0.47\linewidth}
    \includegraphics[width=\linewidth]{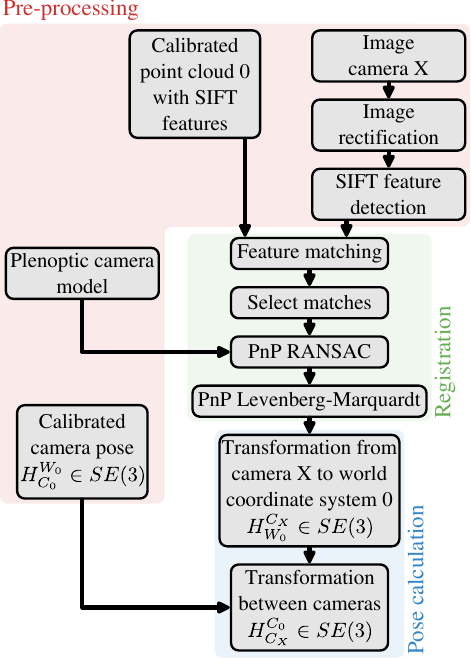}
    \caption{PnP plenoptic multi-camera registration. The reference camera 0 must acquire a short sequence of the scene to obtain an initial calibration and a point cloud. Camera X requires only a single image. The registration is then performed using PnP before calculating the 6-\ac{dof} pose of the camera.}
    \label{fig:pipeline_PnP}
  \end{subfigure}
  \caption{Pipeline of the two proposed camera registration algorithms: the method in Fig.~\ref{fig:pipeline_RANSAC} uses 3D pose estimation via RANSAC, and the method in Fig.~\ref{fig:pipeline_PnP} uses PnP.}
  \label{fig:two_pipelines}
\end{figure}

We propose two benchmark methods for the extrinsic calibration of multiple plenoptic cameras applied to our dataset: one using a 3D transformation estimation via RANSAC and the other a PnP algorithm based on a plenoptic camera model. We use LiFCal~\cite{fleith2024lifcal} to obtain the intrinsic calibration of the cameras and a precisely calibrated point cloud of the environment. The full pipeline is shown in Fig.~\ref{fig:pipeline}, where the multi-camera registration step can use either method.

Extrinsic calibration is performed by moving the cameras within a generic environment. Raw images are processed to generate depth maps and totally focused images from the estimated virtual depth $v$. LiFCal provides intrinsic calibration for each camera, and the resulting depth and calibration data are used to compute relative transformations. Camera 0 is set as the reference, and the other cameras (1 to X) are registered relative to it. In the remainder of the paper, we refer to camera 0 as the reference and camera X as the one to be registered. Homogeneous transformation matrices are denoted by the letter $H$ in $SE(3)$. In the notation $H^{C_0}_{C_X} \in SE(3)$, the superscript indicates the reference frame in which the transformation is expressed (here $C_0$), while the subscript indicates the frame being transformed (here $C_X$). Reference frames are labeled $W$ for the world frame and $C$ for the camera frame. Thus, $H^{C_0}_{C_X} \in SE(3)$ represents the transformation from camera $C_X$ to camera $C_0$ after registration.

\subsection{3D transformation estimation via RANSAC Method}
\label{sec:3DRANSAC}

LiFCal produces an accurate point cloud during intrinsic calibration. Our first method leverages these point clouds and their features using a 3D pose estimation process based on RANSAC, as shown in Fig.~\ref{fig:pipeline_RANSAC}.

SIFT features are extracted from the acquired images and associated with the corresponding 3D points in the point cloud for each camera to be registered. Feature matching between point clouds is performed using a brute-force matcher, and the best matches are retained based on the L2 norm (Euclidean distance), which is well suited for comparing SIFT features~\cite{lowe2004distinctive}.

The matched features are then used in a 3D RANSAC algorithm to align the point clouds and estimate the transformation $H^{W_0}_{W_X} \in SE(3)$ from camera 0’s point cloud to camera X’s. The calibration of camera 0 also provides the transformation from the world frame to the point cloud frame of camera 0, denoted $H^{W_0}_{C_0} \in SE(3)$. Similarly, we can determine $H^{W_X}_{C_X} \in SE(3)$ for camera X. The transformation between camera 0 and camera X is then obtained as:

\begin{align}
	H^{C_0}_{C_X} = H^{W_X}_{C_X} \cdot H^{W_0}_{W_X} \cdot \left( H^{W_0}_{C_0} \right)^{-1} \in SE(3).
	\label{eq:H_C0_CX_3DRANSAC}
\end{align}

\subsection{Plenoptic PnP Method}
\label{sec:PnP}

The 3D RANSAC method (Sec.~\ref{sec:3DRANSAC}) requires point clouds for all cameras, along with motion and intrinsic calibration at each capture. To relax these constraints, we propose the first PnP-based method for plenoptic cameras (see Fig.~\ref{fig:pipeline_PnP}).

Only the point cloud from the first camera’s intrinsic calibration is needed as a reference to estimate the 6-\ac{dof} poses of the other cameras X. SIFT features are extracted and matched to this reference cloud of camera 0.

\begin{figure}[t]
  \centering
  \includegraphics[width=0.99\linewidth]{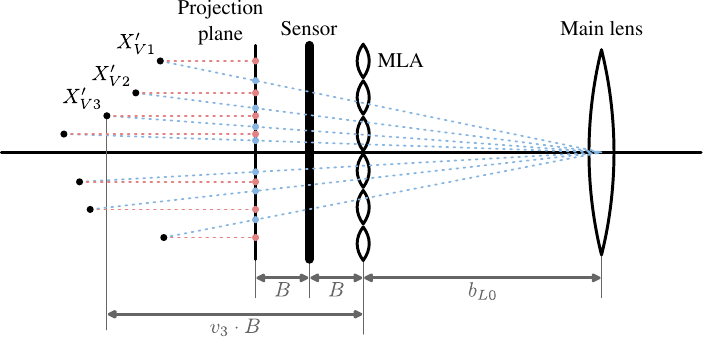}

   \caption{Central perspective projection of the virtual image onto a common image plane. The points are projected along lines through the main lens center (\textit{blue}) onto a plane at a distance of $2B$ from the MLA, instead of using horizontal projection (\textit{orange}). Here, $B$ is the distance between the MLA and the sensor, $b_{L0}$ is the distance between the main lens and the MLA, and $v_3$ is the virtual depth of point $X_{V3}'$.}
   \label{fig:central_perspective_projection}
\end{figure}

For a previously calibrated camera X, a single image is sufficient for registration. We first correct lens distortion and apply a perspective projection using the plenoptic camera model from~\cite{fleith2024lifcal}, which accounts for radial and tangential distortions as well as misalignment between the sensor and the \ac{mla}. In this camera model, $B$ denotes the distance between the \ac{mla} and the camera sensor, and $b_{L0}$ is the distance between main lens and \ac{mla}. The points in the virtual space $X_V'$ are formed at varying distances from the \ac{mla}, defined by the virtual depth $v$ and depending on the object's distance relative to the camera. Instead of being projected horizontally onto the sensor, the points are projected through the main lens center with a projection distance set to $2B$ from the \ac{mla} (corresponding to the maximum measurable depth), as illustrated in Fig.~\ref{fig:central_perspective_projection}. The projected point $X_{proj}=[x_{proj}, y_{proj}]^T$ on the sensor is computed from the virtual point $X_V'=[x_V', y_V', z_V' = v]^T$ via Eq.~(\ref{eq:projX}) and Eq.~(\ref{eq:projY}). SIFT features are then extracted from the corrected image.

\begin{align}
  x_{proj} &= \frac{x_V' - c_x}{v \cdot B + b_{L0}} \cdot (2 \cdot B + b_{L0}) + c_x
  \label{eq:projX}\\
  y_{proj} &= \frac{y_V' - c_y}{v \cdot B + b_{L0}} \cdot (2 \cdot B + b_{L0}) + c_y
  \label{eq:projY}
\end{align}

Features from camera 0’s point cloud and camera X’s image are matched using a brute-force matcher with a k-nearest neighbor method. A cross-check is performed by associating features from the point cloud to the image and vice versa to remove non-mutual matches. The L2 norm is used to retain the best matches.

A plenoptic PnP algorithm is implemented to estimate the pose $H^{C_X}_{W_0} \in SE(3)$ of the view from camera X relative to the point cloud of camera 0. Outliers in the 2D–2D correspondences are first removed using robust fundamental matrix estimation between the reference and query views. An initial estimate is obtained using a RANSAC-based PnP method to filter outliers. The pose is refined by minimizing the reprojection error with non-linear Levenberg-Marquardt minimization scheme. The pose $H^{W_0}_{C_0} \in SE(3)$ of camera 0 with respect to the point cloud is obtained from the intrinsic calibration. The transformation between camera 0 and camera X is then computed as:

\begin{align}
	H^{C_0}_{C_X} = H^{W_0}_{C_X} \cdot \left( H^{W_0}_{C_0} \right)^{-1} \in SE(3).
	\label{eq:H_C0_CX_PnP}
\end{align}

\section{Dataset}
\label{sec:dataset}

\begin{figure}[t]
    \centering

    \begin{tabular}{cccc}
        \begin{subfigure}{0.24\textwidth}
            \includegraphics[width=\linewidth]{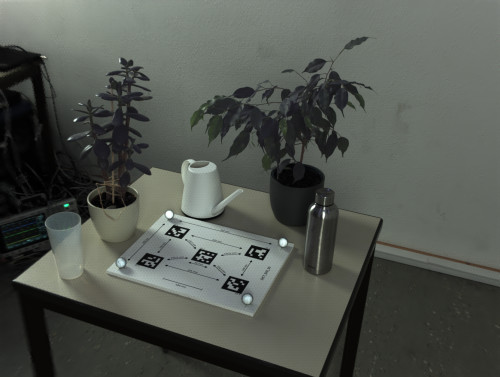}
            \caption{00\_Plants}
        \end{subfigure} &
        \begin{subfigure}{0.24\textwidth}
            \includegraphics[width=\linewidth]{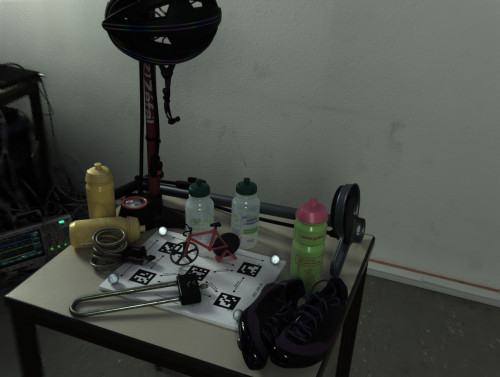}
            \caption{01\_Bike}
        \end{subfigure} &
        \begin{subfigure}{0.24\textwidth}
            \includegraphics[width=\linewidth]{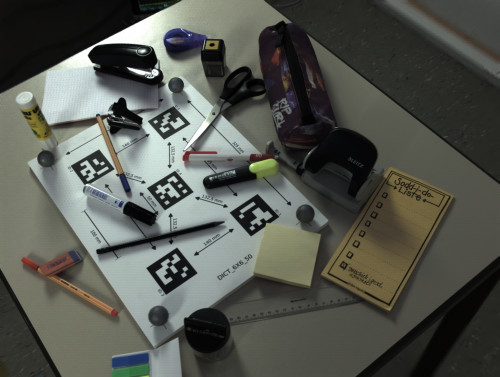}
            \caption{02\_Office}
        \end{subfigure} &
        \begin{subfigure}{0.24\textwidth}
            \includegraphics[width=\linewidth]{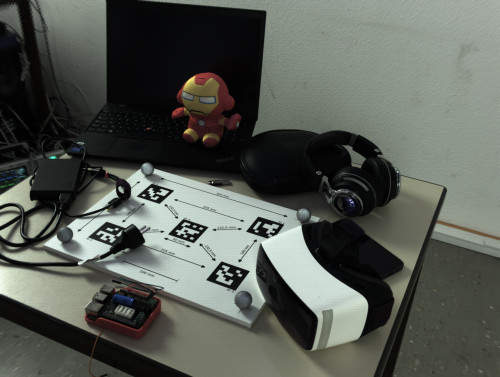}
            \caption{03\_Electronics}
        \end{subfigure}
    \end{tabular}

    \vspace{1em}

    \begin{tabular}{ccc}
        \begin{subfigure}{0.24\textwidth}
            \includegraphics[width=\linewidth]{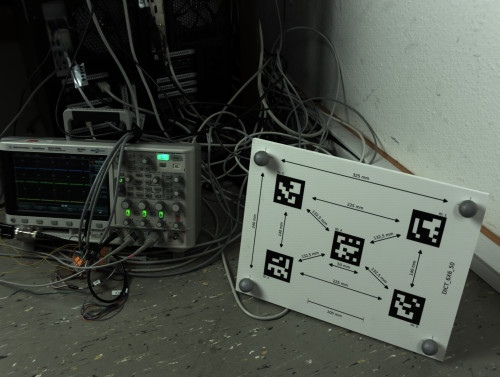}
            \caption{04\_Oscilloscope}
        \end{subfigure} &
        \begin{subfigure}{0.24\textwidth}
            \includegraphics[width=\linewidth]{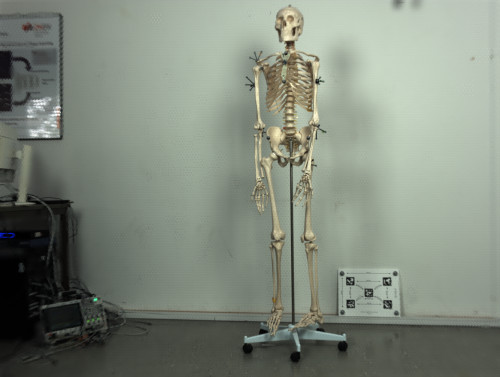}
            \caption{05\_Skeleton}
        \end{subfigure} &
        \begin{subfigure}{0.24\textwidth}
            \includegraphics[width=\linewidth]{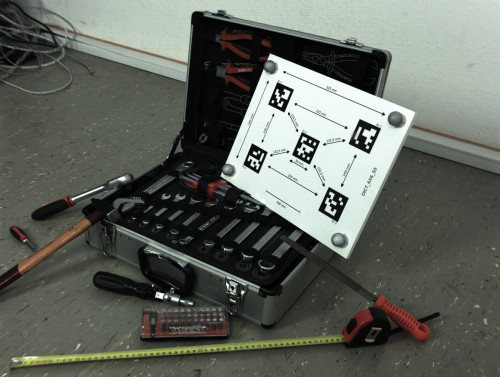}
            \caption{06\_Tools}
        \end{subfigure}
    \end{tabular}

    \caption{Totally focused images processed from sample views of the scenes in the dataset.}
    \label{fig:images_dataset}
\end{figure}

We present a dataset of synchronized sequences from two high-resolution Raytrix R32 plenoptic cameras with Vicon-based 6~\ac{dof} ground truth. Designed for multi-camera registration, it also supports applications such as \ac{slam}, \ac{sfm}, and \ac{nvs}. LiFMCR overcomes the limitations of existing datasets, as summarized in Table~\ref{tab:datasets}, and is the first to provide synchronized sequences from multiple focused plenoptic cameras with accurate ground truth. It comprises seven distinct scenes (see Fig.~\ref{fig:images_dataset} and supplementary material for trajectory plots), including raw images from both cameras, \ac{mla} calibration data, reference marker, and ground truth poses. See the supplementary material for a detailed overview of the dataset structure and content. 

\renewcommand\tabularxcolumn[1]{>{\centering\arraybackslash}m{#1}} 
\newcolumntype{Y}{>{\centering\arraybackslash}X} 

\begin{table}[htbp]
\caption{Comparison of existing main plenoptic camera datasets with our LiFMCR dataset.}
\label{tab:datasets}
\centering
\begin{tabularx}{\textwidth}{|Y|c|c|c|c|c|}
\hline
\textbf{Dataset} & \textbf{Year} & \textbf{\makecell[c]{Camera\\type}} & \textbf{\makecell[c]{Multi-\\camera}} & \textbf{\makecell[c]{Ground\\truth}} \\
\hline
LiFMCR (our dataset) & 2025 & Raytrix R32 & Yes & Yes \\
\hline
The Stanford Multiview Light Field Datasets~\cite{dansereau2019liff} & 2019 & Lytro Illum & Yes & No \\
\hline
Stanford Lytro Light Field Archive~\cite{raj2016stanfordLF2016} & 2016 & Lytro Illum & No & No \\
\hline
4D Light Field Dataset~\cite{honauer2016dataset} & 2016 & Blender & No & Synthetic \\
\hline
Light-Field Image Dataset~\cite{rerabek2016new} & 2016 & Lytro Illum & No & No \\
\hline
Light field Saliency Dataset (LFSD)~\cite{li2014saliency} & 2014 & Lytro & No & No \\
\hline
A 4D Light-Field Dataset and CNN Architectures for Material Recognition~\cite{wang20164d} & 2016 & Lytro Illum & No & No \\
\hline
LCAV-31: A Dataset for Light Field Object Recognition~\cite{afonso2013lcav} & 2013 & Lytro & No & No \\
\hline
\end{tabularx}
\end{table}

\begin{figure}[htbp]
    \centering
    \begin{minipage}[b]{0.31\linewidth}
        \centering
        \includegraphics[width=\linewidth]{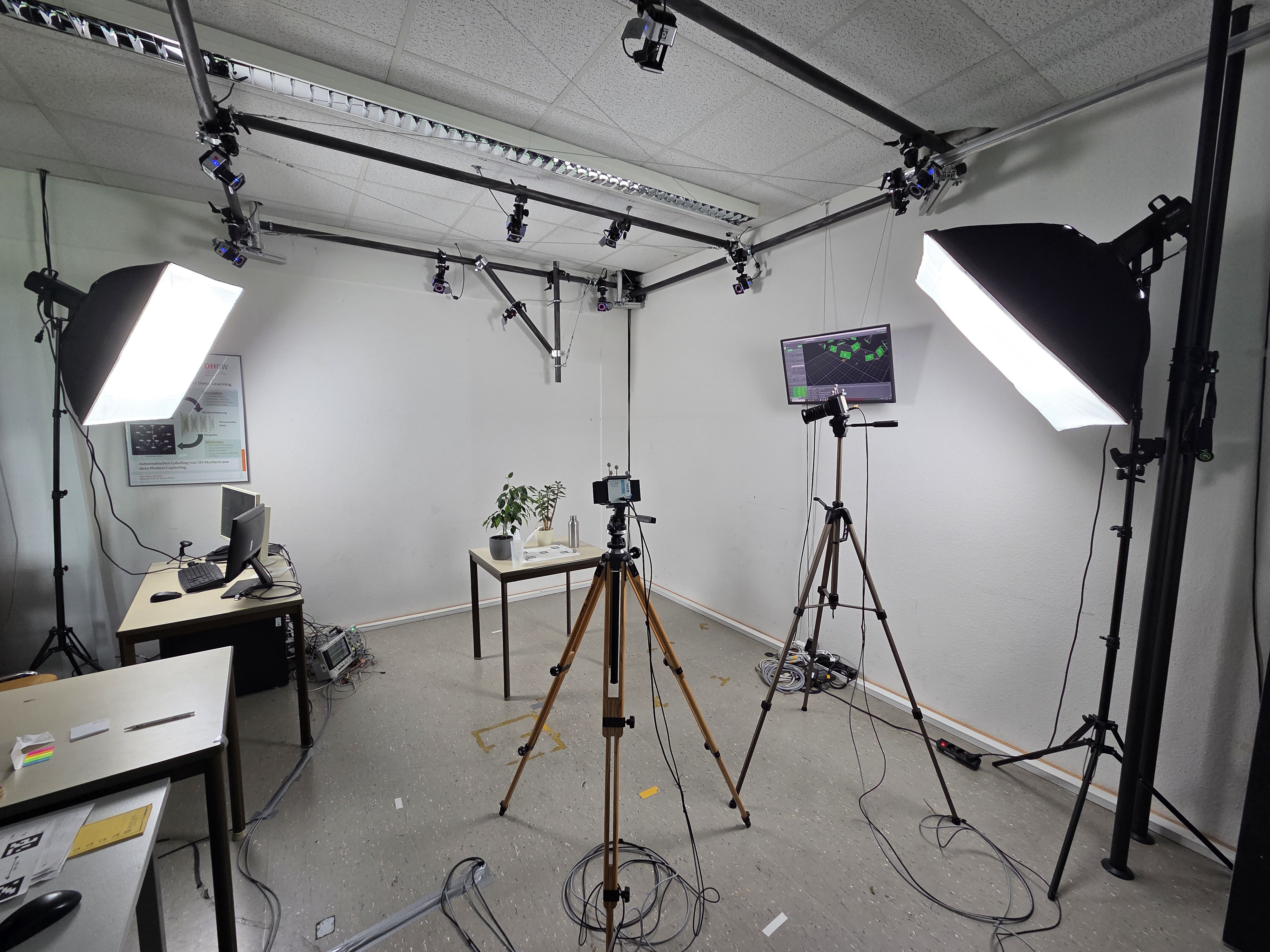}
        \caption{Synchronized acquisition setup using plenoptic cameras and the Vicon motion tracking system.}
        \label{fig:setup}
    \end{minipage}
    \hfill
    \begin{minipage}[b]{0.31\linewidth}
        \centering
        \includegraphics[width=\linewidth]{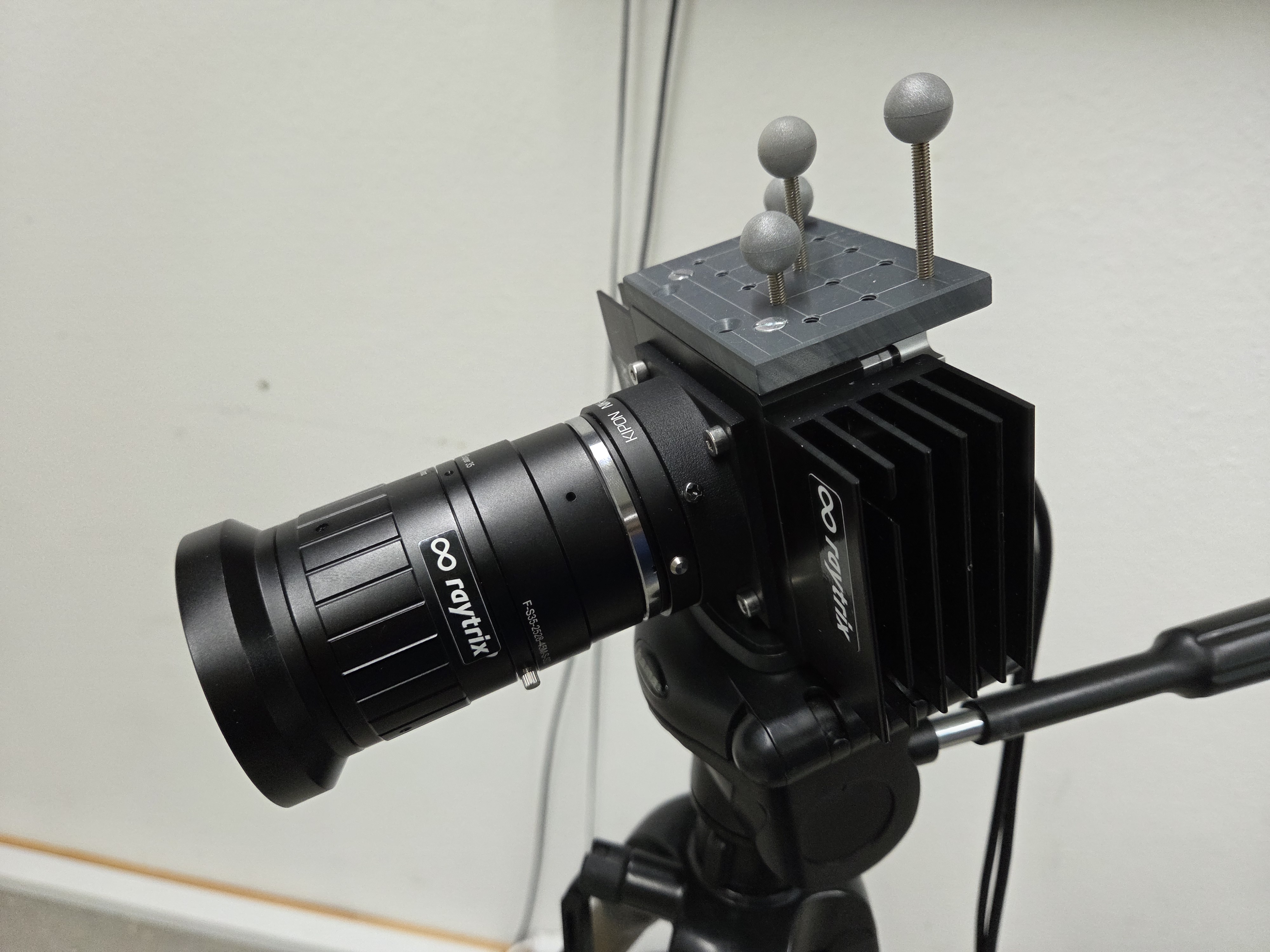}
        \caption{Marker setup with four infrared reflective spheres for unique identification and 6-\ac{dof} tracking.}
        \label{fig:camera_marker}
    \end{minipage}
    \hfill
    \begin{minipage}[b]{0.31\linewidth}
        \centering
        \includegraphics[width=\linewidth]{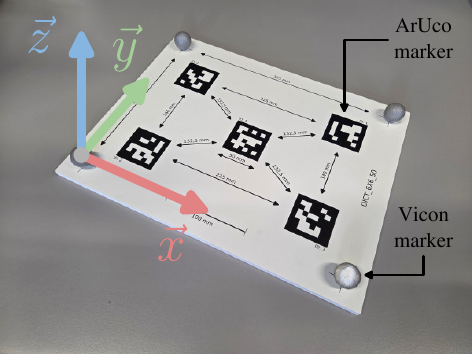}        
        \caption{Plate with Vicon (defining the reference frame) and ArUco (providing scale for intrinsic calibration) markers.}
        \label{fig:marker_plate}
    \end{minipage}
\end{figure}

\subsection{Camera Specifications}

The Raytrix R32 cameras are built on a Basler boost r boa6500-36cc body with a global shutter sensor XGS32000 by Onsime. For a good trade-off between field of view and angular resolution of the captured light field, they are equipped with a Basler main lens F-S35-2528-45M-S-SD with a focal length of $f_L = 25$~mm. Both cameras share identical specifications, which are summarized in Table~\ref{tab:cameraData}. 

\begin{table}[htbp]
    \centering
    \begin{minipage}[t]{0.5\textwidth}
        \vspace{0pt} 
        \centering
        \caption{Specifications for both Raytrix cameras used for dataset acquisition.}
        \label{tab:cameraData}
        \begin{tabular}{|c|c|}
        \hline
        \textbf{Specification} & \textbf{Plenoptic camera value} \\
        \hline
        Camera model & Raytrix R32 \\
        Pixel size & 3.2~µm \\
        Resolution & 6560 $\times$ 4948 pixels \\
        Focal length & 25~mm \\
        Aperture & 1:2.4 \\
        Color channels & 3 \\
        \hline
        \end{tabular}
    \end{minipage}
    \hfill
    \begin{minipage}[t]{0.45\textwidth}
        \vspace{0pt} 
        \centering
        \includegraphics[width=0.8\linewidth]{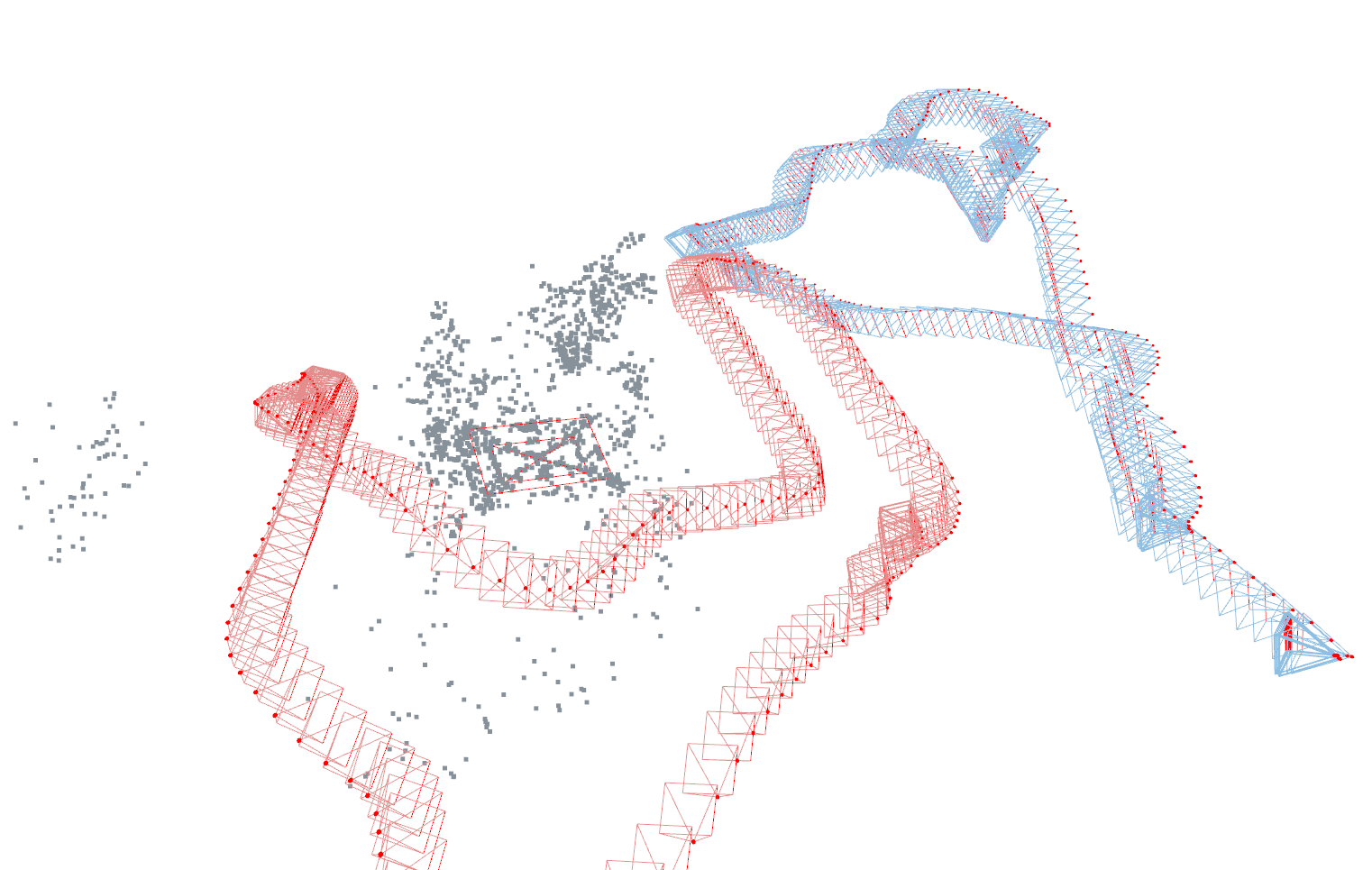}
        \caption{Tracking of cameras and plate positions in the 00\_Plants scene.}
        \label{fig:tracking_cameras}
    \end{minipage}
\end{table}

\subsection{Ground Truth Acquisition}

The dataset includes ground truth 6-\ac{dof} poses, acquired using the Vicon system~\cite{vicon}, an optical motion capture system employing thirteen infrared cameras. It tracks the 3D positions of reflective markers with sub-millimeter accuracy via triangulation, enabled by precise calibration and synchronization. Fig.~\ref{fig:setup} shows the full acquisition setup.

Each plenoptic camera carries a unique four-marker plate (Fig.~\ref{fig:camera_marker}) for 6-\ac{dof} pose tracking. The scenes also include an ArUco plate (Fig.~\ref{fig:marker_plate}) for metric scene scale, as introduced in LiFCal~\cite{fleith2024lifcal}. It is also intended for future applications using the dataset. Vicon markers at its corners enable pose tracking. When placed on the table, this plate defines the world frame: origin at the bottom-left marker, $x$ along the table’s length (rightward), $y$ along its width, and $z$ upward.

With markers on both cameras and the plate, their 6-\ac{dof} poses are tracked at 80~Hz. The reference frame defined by the plate remains fixed in the environment, allowing the plate to be tracked even when it is moved in the scene. Fig.~\ref{fig:tracking_cameras} shows pose tracking for the cameras and the plate in scene 00\_Plants.

\subsection{Aquisiton Sytem}

Vicon’s Tracker records ground truth data at 80~Hz on a separate system. The Vicon Lock Lab output is downsampled by a factor of 8 to 10~Hz to trigger both cameras, ensuring precise alignment between image acquisition and motion capture data. This also keeps the data rate manageable at around 1.9~GB/s for the two cameras. Raw image data is captured using the software RxLive by Raytrix GmbH on an Intel\textsuperscript{\tiny\textregistered} Core\textsuperscript{\tiny TM} i9-10980XE × 36 system with 128~GB RAM.

\section{Evaluation}
\label{sec:evaluation}

\newcolumntype{L}[1]{>{\raggedright\arraybackslash}p{#1}} 
\newcolumntype{C}[1]{>{\centering\arraybackslash}p{#1}}  

\begin{table}[t]
\caption{Relative evaluation of the translation error (T.error) and rotation error (R. error) of the two benchmark methods compared to the ground truth. The lowest value in each column is shown in \textbf{bold} and the highest value is \underline{underlined} (lower is better).}
\label{tab:resultsRelative}
\centering
\begin{tabular}{|L{2.3cm}|*{8}{C{1.1cm}|}}
\hline
\multirow{2}{*}{Sequence} & \multicolumn{4}{c|}{3D RANSAC Method} & \multicolumn{4}{c|}{Plenoptic PnP Method} \\
& \multicolumn{2}{c|}{T. error [mm]} & \multicolumn{2}{c|}{R. error [°]} & \multicolumn{2}{c|}{T. error [mm]} & \multicolumn{2}{c|}{R. error [°]} \\
& RMSE & SD & RMSE & SD & RMSE & SD & RMSE & SD \\
\hline
00\_Plants & 49.73 & 23.99 & \textbf{1.62} & \textbf{0.71} & 51.45 & 23.81 & \textbf{1.63} & \textbf{0.69} \\
01\_Bike & 61.40 & 24.08 & 2.04 & 0.76 & 60.24 & 22.81 & 2.02 & 0.77 \\
02\_Office & \textbf{31.11} & 17.44 & 2.21 & 1.03 & \textbf{31.53} & 18.12 & 2.20 & 1.03 \\
03\_Electronics & 46.10 & \textbf{13.54} & 2.04 & 0.76 & 44.14 & \textbf{15.35} & 2.07 & 0.96 \\
04\_Oscilloscope & 48.89 & 24.83 & 2.30 & 0.82 & 47.15 & 25.32 & 2.39 & 0.70 \\
05\_Skeleton & \underline{69.06} & \underline{28.20} & \underline{3.76} & \underline{1.53} & \underline{67.41} & \underline{26.88} & \underline{3.76} & \underline{1.53} \\
06\_Tools & 42.51 & 18.03 & 1.99 & 1.03 & 42.19 & 16.60 & 2.04 & 0.97 \\
\hline
\end{tabular}
\end{table}

\begin{table}[t]
\caption{Absolute evaluation of the translation error (T.error) and rotation error (R. error) of the two benchmark methods compared to the ground truth. The lowest value in each column is shown in \textbf{bold} and the highest value is \underline{underlined} (lower is better).}
\label{tab:resultsAbsolute}
\centering
\begin{tabular}{|L{2.3cm}|*{8}{C{1.1cm}|}}
\hline
\multirow{2}{*}{Sequence} & \multicolumn{4}{c|}{3D RANSAC Method} & \multicolumn{4}{c|}{Plenoptic PnP Method} \\
& \multicolumn{2}{c|}{T. error [mm]} & \multicolumn{2}{c|}{R. error [°]} & \multicolumn{2}{c|}{T. error [mm]} & \multicolumn{2}{c|}{R. error [°]} \\
& RMSE & SD & RMSE & SD & RMSE & SD & RMSE & SD \\
\hline
00\_Plants & 209.26 & 18.52 & 7.82 & 3.78 & 207.04 & 23.10 & \underline{8.06} & \underline{3.71} \\
01\_Bike & \underline{214.83} & 26.11 & 7.94 & \underline{3.90} & \underline{214.74} & 26.56 & \underline{8.06} & 3.63 \\
02\_Office & 200.37 & 22.60 & \underline{8.15} & 3.19 & 201.68 & 28.32 & 7.90 & 3.51 \\
03\_Electronics & 207.31 & \textbf{16.02} & 7.19 & 3.04 & 211.73 & \textbf{20.10} & 7.84 & 3.43 \\
04\_Oscilloscope & 134.17 & 29.11 & 7.96 & 3.00 & 133.38 & 29.92 & 8.02 & 3.10 \\
05\_Skeleton & 158.82 & 27.14 & \textbf{6.95} & \textbf{2.95} & 155.08 & 28.18 & 7.16 & 2.94 \\
06\_Tools & \textbf{117.31} & \underline{32.21} & 6.97 & 3.07 & \textbf{105.35} & \underline{30.13} & \textbf{6.36} & \textbf{2.67} \\
\hline
\end{tabular}
\end{table}

We evaluated two benchmark registration methods on our dataset LiFMCR. The 3D RANSAC (Sec.~\ref{sec:3DRANSAC}) and plenoptic PnP (Sec.~\ref{sec:PnP}) algorithms were evaluated by comparing the results with the ground truth data from the Vicon system. Additional experiments are provided in the supplementary material.

\subsection{Experiments description}

To evaluate performance, images were extracted at fixed intervals from all sequences. 6-\ac{dof} camera registration was performed for both cameras, with each camera used subsequently as source and target. The measured data and the ground truth data (from the Vicon system) were aligned based on the reference marker plate (Fig.~\ref{fig:marker_plate}). See the supplementary material for more details.

\subsection{Experiments results}

A first experiment compares the relative pose difference between consecutive frames with Vicon ground truth (Table~\ref{tab:resultsRelative}). The rotation \ac{rmse} is low for both methods, mostly around 2°, with a \ac{sd} of about 1° or less, indicating consistency. The translation \ac{rmse} remains around 50~mm across all scenes, providing a strong reference for this dataset.

We then assess absolute pose error against ground truth (Table~\ref{tab:resultsAbsolute}). Although the rotation \ac{rmse} is higher (between 6.95° and 8.15° for RANSAC and between 6.36° and 8.06° for PnP), its low \ac{sd} supports the methods’ validity. Note the relatively larger absolute \ac{rmse}, which is nonetheless highly consistent (\ac{sd} bellow 30~mm). This suggests a systematic offset, likely due to the uncorrected shift between the camera’s optical center and the Vicon markers. This is highlighted by examining the translation error separately along each axis (see supplementary material). The resulting offset is consistent with a plausible shift of the optical center, located close to the principal axis of the main lens.

\section{Conclusion}
\label{sec:conclusion}

We introduced a new dataset to advance research in plenoptic camera registration and multi-view reconstruction. It provides synchronized, high-resolution sequences from multiple \ac{mla}-based light field cameras, paired with accurate 6-\ac{dof} ground truth poses provided by a Vicon motion capture system. This unique combination of data makes it a valuable dataset for tasks such as calibration, pose estimation, \ac{nvs}, and scene understanding.

To demonstrate the utility of this dataset, we proposed two benchmark methods: a 3D pose estimation via RANSAC point cloud alignment and a plenoptic PnP algorithm, both designed based on a plenoptic camera model. Experimental results show strong agreement with ground truth, highlighting the dataset's relevance for future work in light field reconstruction, \ac{slam}, and related applications.

%
%
%
\bibliographystyle{splncs04}
\bibliography{main}
%


\clearpage

\title{LiFMCR: Dataset and Benchmark for Light Field Multi-Camera Registration\\Supplementary Material}
\titlerunning{LiFMCR}
%

\makeatother
\author{Aymeric Fleith\thanks{These authors contributed equally.\protect\label{XX}}\inst{,1,2} \and
Julian Zirbel\repthanks{X}\inst{,1,2} \and 
Daniel Cremers\inst{1} \and
Niclas Zeller\inst{2}}

\authorrunning{Fleith and Zirbel et al.}
%
\institute{Technical University of Munich, Munich, Germany\\
\email{\{aymeric.fleith, julian.zirbel, cremers\}@tum.de}\\
\and Karlsruhe University of Applied Sciences, Karlsruhe, Germany\\
\email{niclas.zeller@h-ka.de}}

\maketitle              

\renewcommand*{\thesection}{\Alph{section}}
\renewcommand\tabularxcolumn[1]{>{\centering\arraybackslash}m{#1}} 

\section{Introduction}
\label{sec:SM_introduction}

This supplementary material provides additional details beyond those in the main paper.
More specifically, this document presents more precisely the structure of the dataset in Sec.~\ref{sec:SM_DatasetStructure}, and its content, including the types of sequences and the associated number of frames in Sec.~\ref{sec:SM_DatasetContent}. A graphical representation of the trajectories of the two cameras in the main sequence of each scene is shown in Sec.~\ref{sec:SM_CameraTrajectories}. The structure of the \ac{mla} calibration file provided with the data is explained in Sec.~\ref{sec:SM_MLACalibrationFile}. Sec.~\ref{sec:SM_DataAlignment} elaborates on the data alignment used for evaluation. Additional evaluations of the two benchmark methods are presented in Sec.~\ref{sec:SM_EvaluationdDifference} for comparison. Finally, Sec.~\ref{sec:SM_systematicOffset} highlights the origin of a systematic offset in absolute translation errors.

\section{Dataset Structure}
\label{sec:SM_DatasetStructure}

The dataset follows a hierarchical folder structure. At the highest level, each scene is stored in its own directory, which contains subfolders for individual sequences. Each sequence folder contains the data summarized in Table~\ref{tab:datasetStructure}. The calibration folder, located alongside the scene folders, follows the structure given in Table~\ref{tab:datasetStructureCalibration}.

\begin{table}[htbp]
\caption{Subfolder structure within each sequence directory of the LiFMCR dataset.}
\label{tab:datasetStructure}
\centering
\begin{tabularx}{\textwidth}{|c|Y|c|}
\hline
\textbf{Folder}&\textbf{Name} & \textbf{Description} \\
\hline
Vicon & sequence\_XX.csv    & Trajectories of tracked objects \\
TypeE\_40398673 & TypeE\_40398673\_XX\_Raw.bmp     & Raw plenoptic camera frames  \\
TypeE\_40398678 & TypeE\_40398678\_XX\_Raw.bmp & Raw plenoptic camera frames \\
\hline
\end{tabularx}
\end{table}

\begin{table}[htbp]
\caption{Subfolder structure within the calibration directory of the LiFMCR dataset.}
\label{tab:datasetStructureCalibration}
\centering
\begin{tabularx}{\textwidth}{|c|Y|c|}
\hline
\textbf{Folder}&\textbf{Name} & \textbf{Description} \\
\hline
01\_White\_Images & TypeE\_4039867X\_XX\_Raw.bmp    & Raw white images \\
02\_MLA\_Calibration & TypeE\_4039867X\_MLA.xml     & MLA calibration file  \\
03\_Vicon & Object.vsk & Vicon marker clusters \\
\hline
\end{tabularx}
\end{table}

\subsubsection{Vicon}
This folder contains a \texttt{.csv} file with the trajectories of three tracked objects: the two cameras and a reference plane of known geometry with an ArUco marker pattern.

\subsubsection{Raytrix}
Each camera has its own directory with raw and preprocessed plenoptic frames. In our setup, the cameras are labeled 
\texttt{TypeE\_40398678} (referred to as \texttt{cam0} in the Vicon file) and 
\texttt{TypeE\_40398673} (referred to as \texttt{cam2} in the Vicon file). The raw frames are stored in \texttt{.bmp} format.

\subsubsection{Calibration}
The file TypeE\_4039867x\_MLA.xml contains the calibration data for the \ac{mla}. It is explained in detail in Sec.~\ref{sec:SM_MLACalibrationFile}.

\section{Dataset Content}
\label{sec:SM_DatasetContent}

The dataset consists of seven different indoor scenes captured by two synchronized Raytrix R32 plenoptic cameras. All scenes include a sequence in which the two cameras perform random movements in front of the scene. These are considered the main sequences of the dataset. In addition, some scenes have additional sequences in which certain parameters vary. The different sequences of the dataset and the associated number of images (identical for both cameras) are summarized in Table~\ref{tab:datasetContent}.

\begin{table}[htbp]
\caption{Content of the LiFMCR dataset.}
\label{tab:datasetContent}
\centering
\begin{tabularx}{\textwidth}{|c|Y|c|}
\hline
\textbf{Scene} & \textbf{Sequence type} & \textbf{Number of frames} \\
\hline
\multirow{2}{*}{00\_Plants} & Random camera movements & 323 \\
& Handheld movements around the scene & 632 \\
\hline
\multirow{3}{*}{01\_Bike}  & Random camera movements & 434 \\
& Handheld movements around the scene & 781 \\
& Movements in $x$, $y$, $z$ directions & 250 \\
\hline
\multirow{3}{*}{02\_Office} & Random camera movements & 328 \\
& Handheld movements around the scene & 688 \\
& Fast movements & 247 \\
\hline
\multirow{3}{*}{03\_Electronics} & Random camera movements & 323 \\
& Handheld movements around the scene & 185 \\
& Random movements with lower exposure & 428 \\
\hline
\multirow{1}{*}{04\_Oscilloscope} & Random camera movements & 406 \\
\hline
\multirow{3}{*}{05\_Skeleton} & Random camera movements & 533 \\
& Handheld close movements & 758 \\
& Cameras in circle & 677 \\
\hline
\multirow{1}{*}{06\_Tools} & Random camera movements & 471 \\
\hline
\end{tabularx}
\end{table}

\section{Camera Trajectories in Sequences}
\label{sec:SM_CameraTrajectories}

\begin{figure}[htbp]
    \centering

    \begin{tabular}{cc}
        \begin{subfigure}{0.48\textwidth}
            \includegraphics[width=0.97\linewidth]{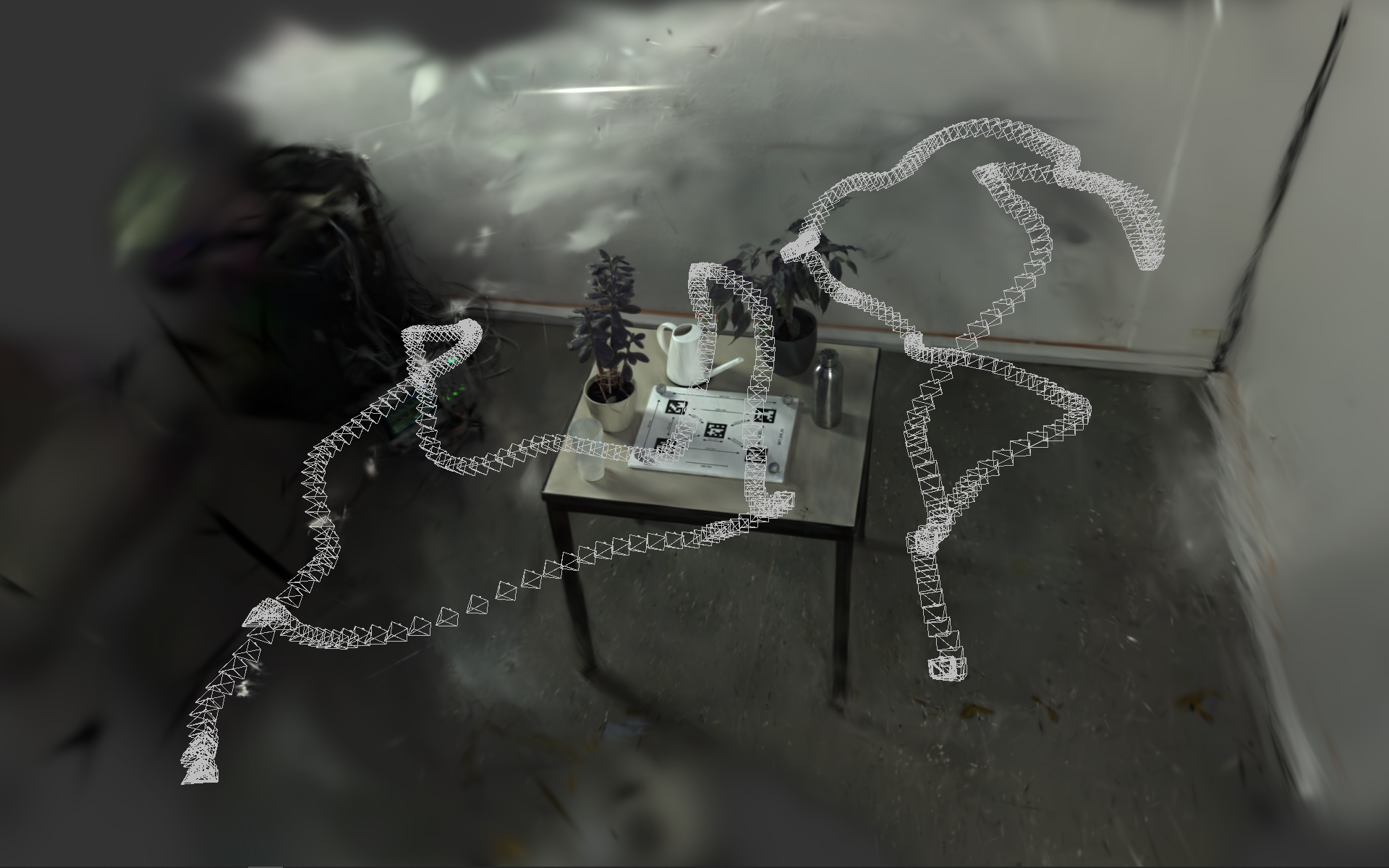}
            \caption{00\_Plants}
        \end{subfigure} &
        \begin{subfigure}{0.48\textwidth}
            \includegraphics[width=0.97\linewidth]{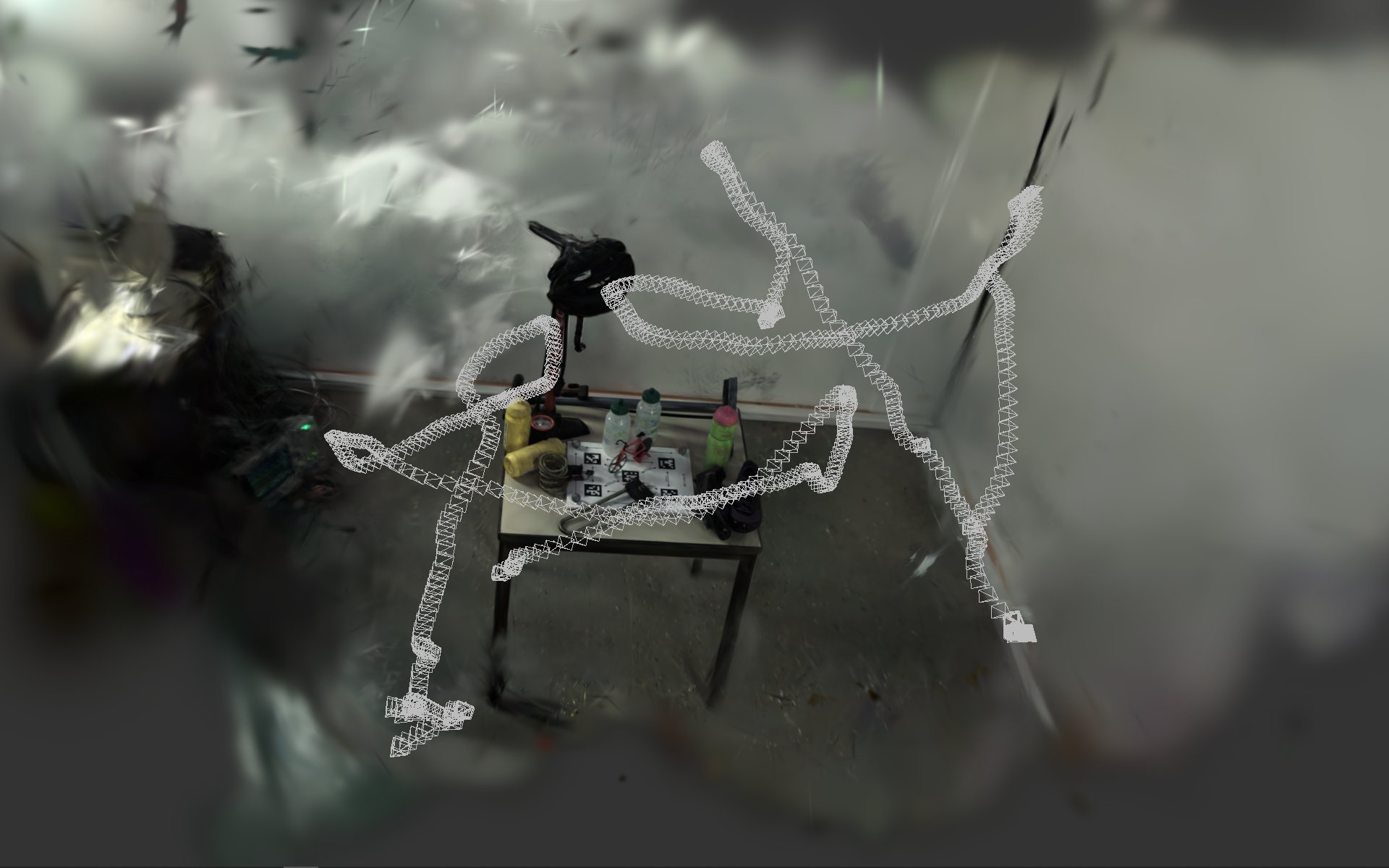}
            \caption{01\_Bike}
        \end{subfigure}
    \end{tabular}

    \vspace{1em}

    \begin{tabular}{cc}
        \begin{subfigure}{0.48\textwidth}
            \includegraphics[width=0.97\linewidth]{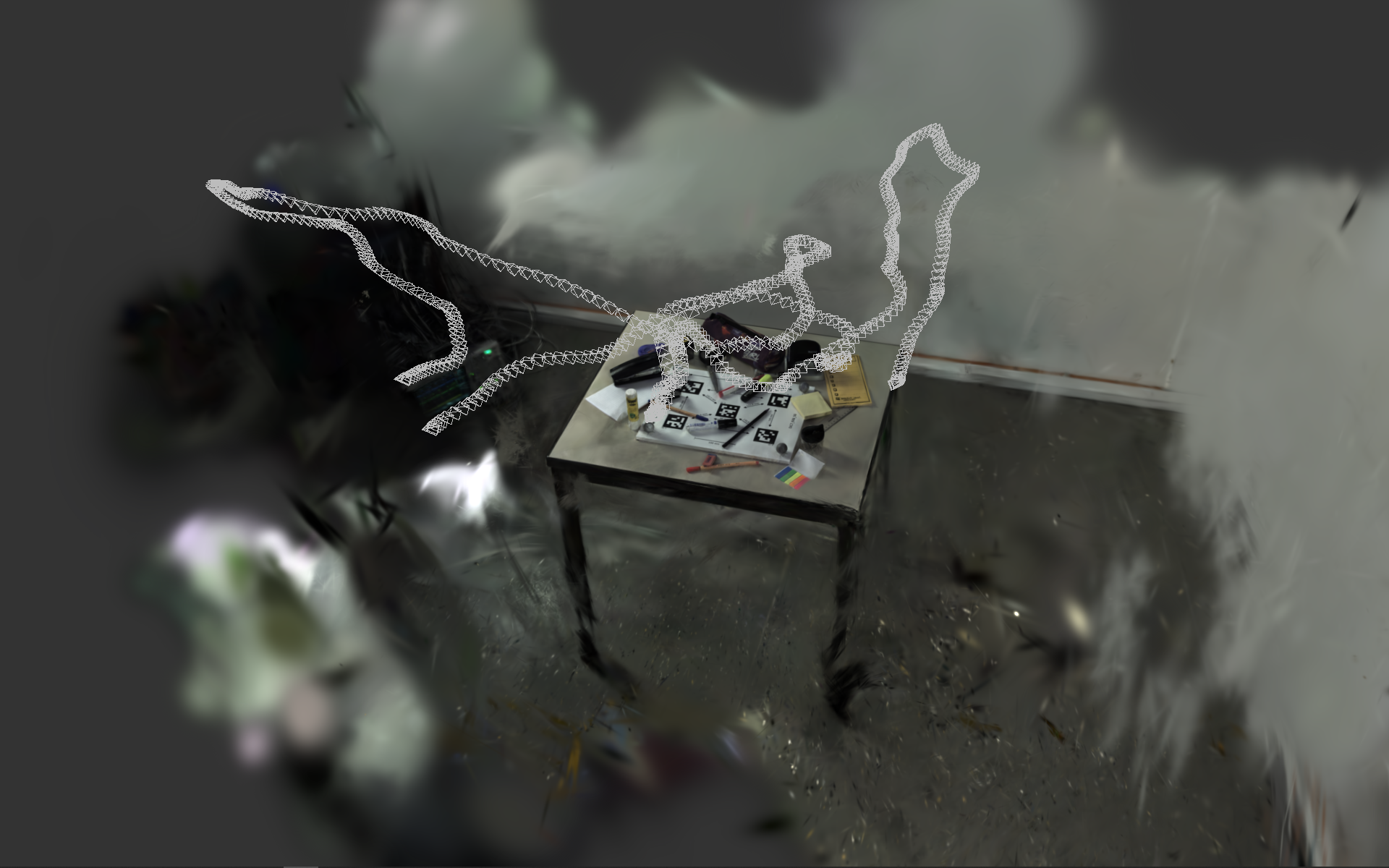}
            \caption{02\_Office}
        \end{subfigure} &
        \begin{subfigure}{0.48\textwidth}
            \includegraphics[width=0.97\linewidth]{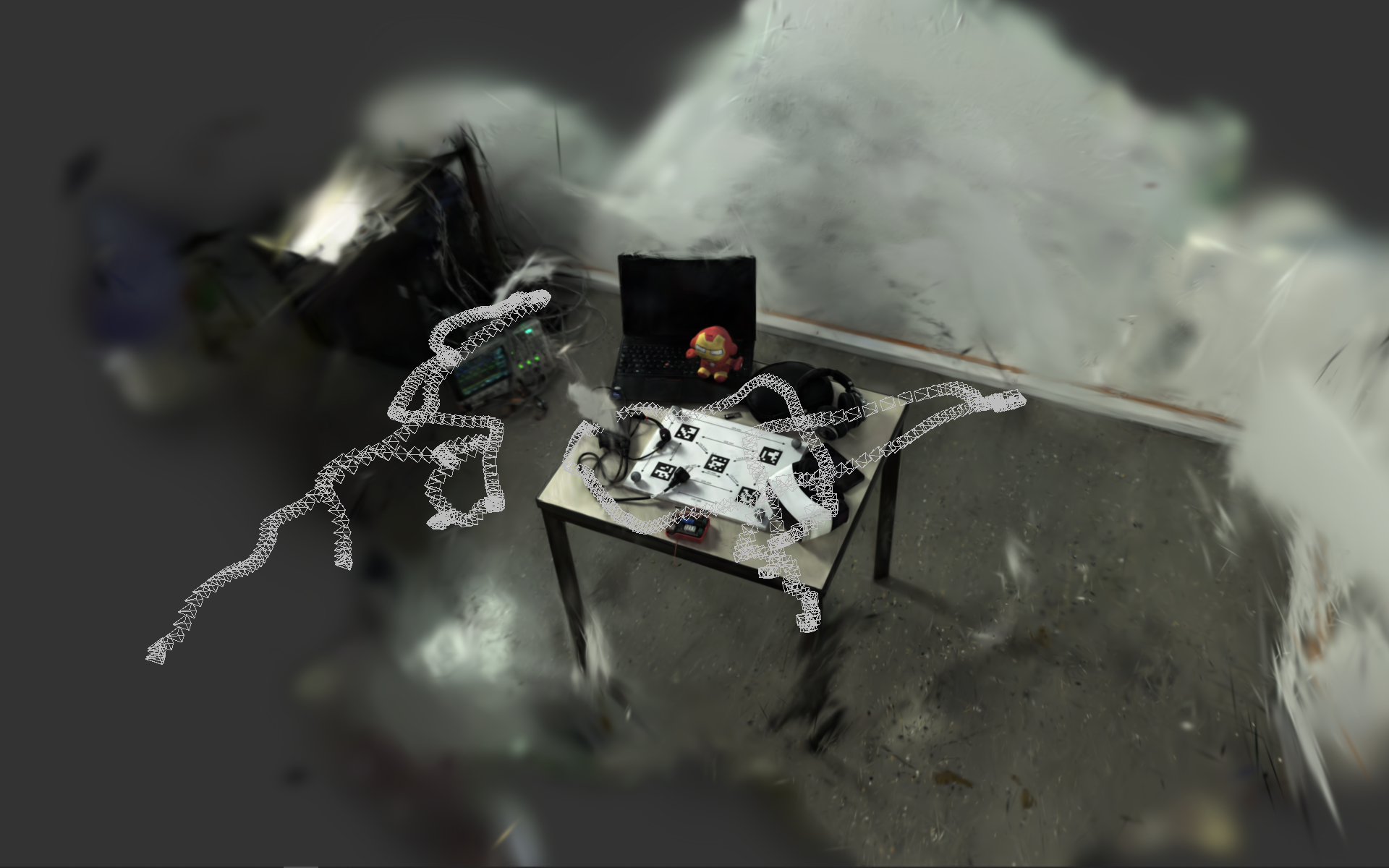}
            \caption{03\_Electronics}
        \end{subfigure}
    \end{tabular}

    \vspace{1em}

    \begin{tabular}{cc}
        \begin{subfigure}{0.48\textwidth}
            \includegraphics[width=0.97\linewidth]{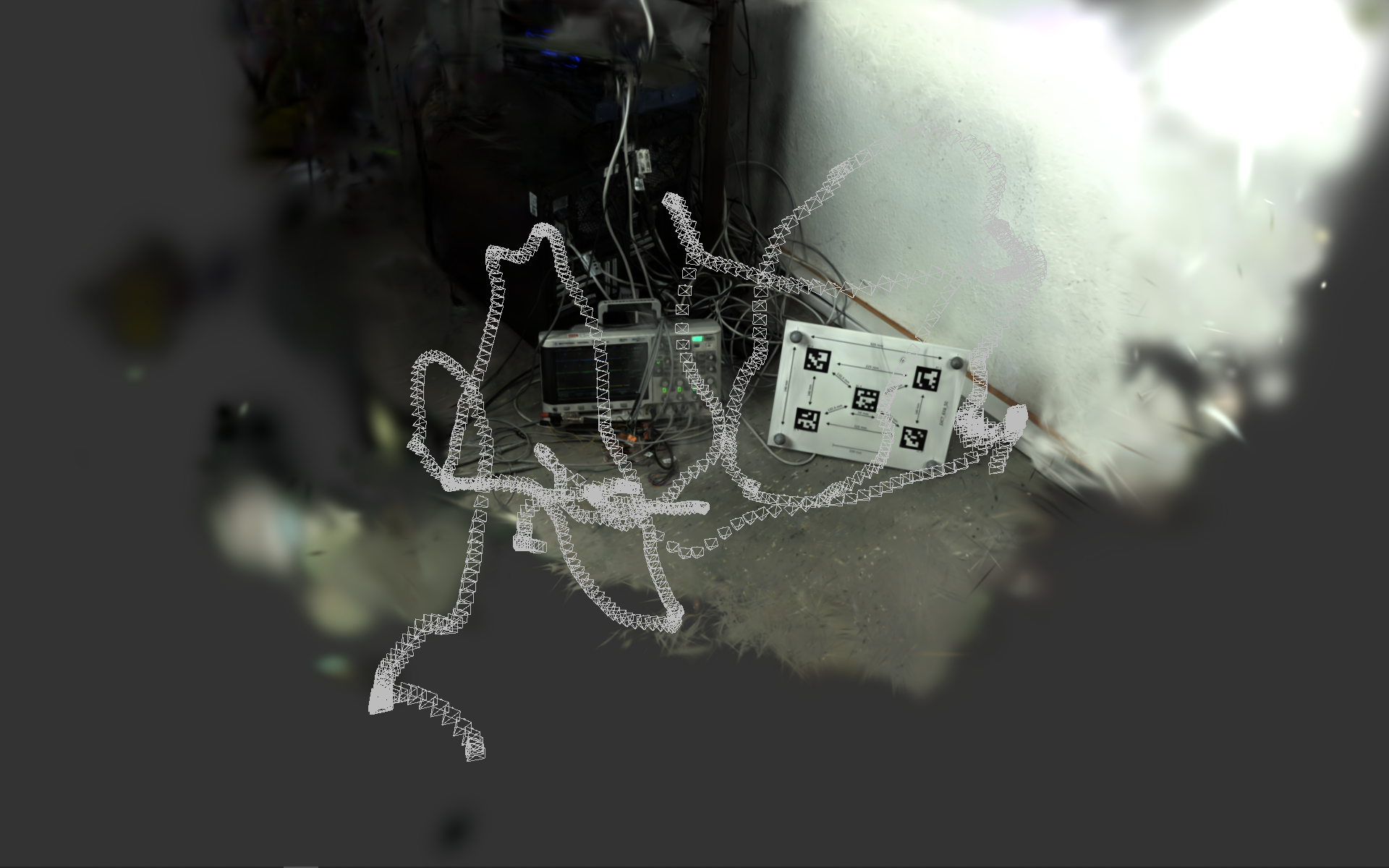}
            \caption{04\_Oscilloscope}
        \end{subfigure} &
        \begin{subfigure}{0.48\textwidth}
            \includegraphics[width=0.97\linewidth]{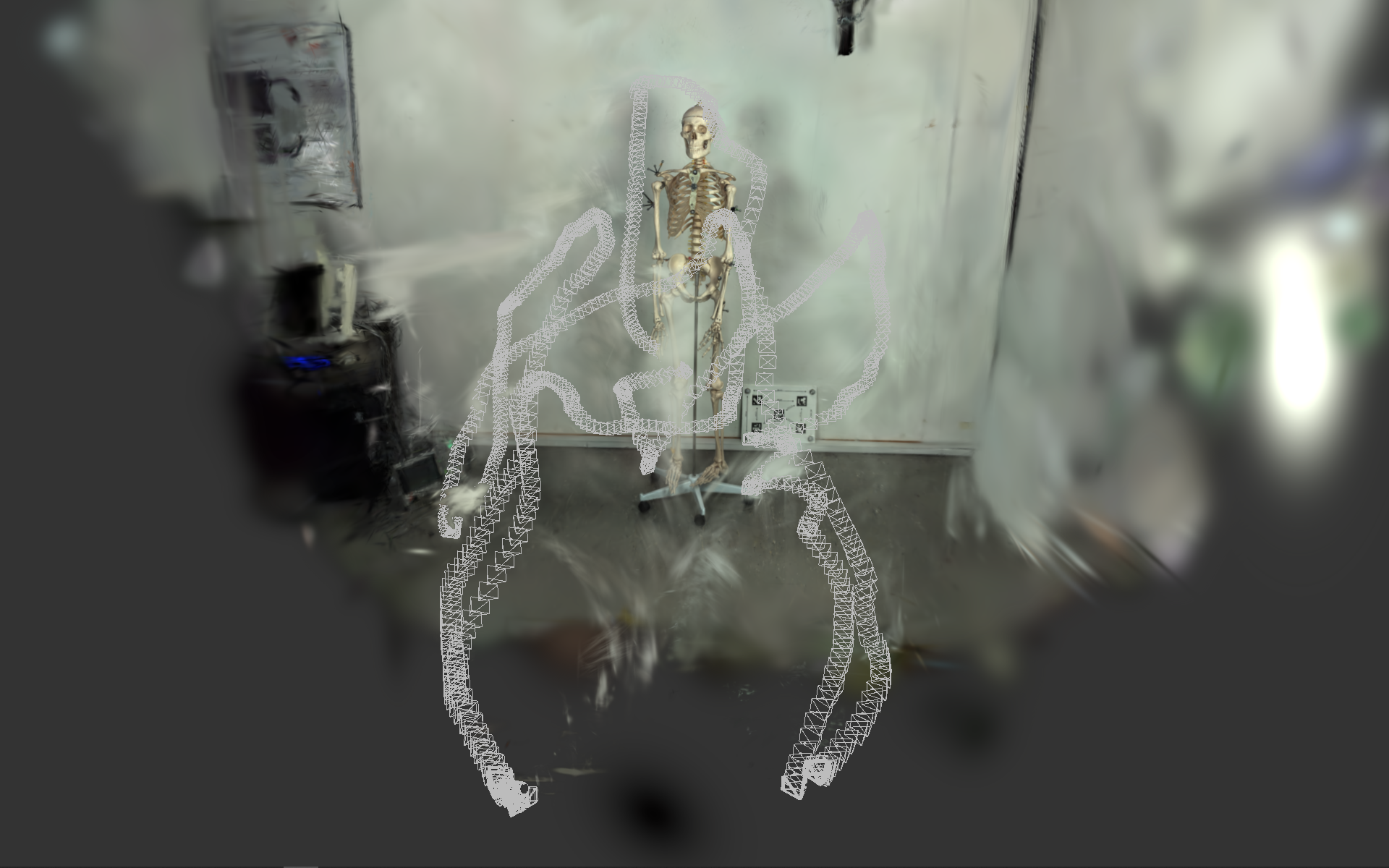}
            \caption{05\_Skeleton}
        \end{subfigure}
    \end{tabular}

    \vspace{1em}

    \begin{tabular}{c}
        \begin{subfigure}{0.48\textwidth}
            \includegraphics[width=0.97\linewidth]{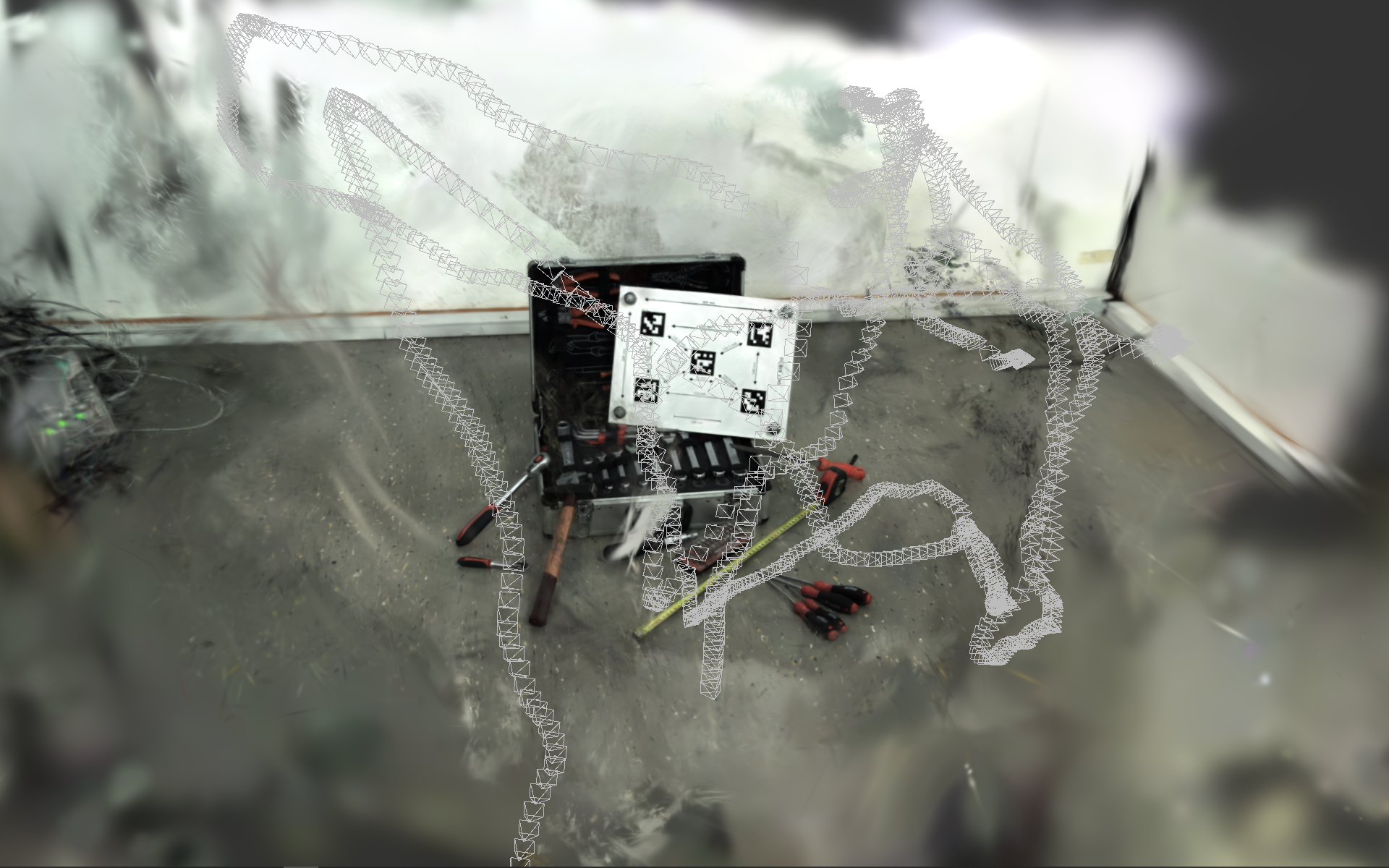}
            \caption{06\_Tools}
        \end{subfigure}
    \end{tabular}

    \caption{Graphical representation of camera trajectories and frame poses for the two Raytrix R32 cameras in the main sequences of each scene.}
    \label{fig:CameraTrajectories}
\end{figure}

In the main sequence of each scene, the two Raytrix R32 cameras move randomly in front of the scene as explained in Sec.~\ref{sec:SM_DatasetContent}. Fig.~\ref{fig:CameraTrajectories} shows the trajectories of the cameras as well as the number of images in the main sequences by graphically displaying the poses of the views. The visualization is performed using Jawset Postshot~\cite{Postshot}, an end-to-end software for radiance fields.
This visualization also demonstrates the potential of the dataset, particularly through the application of radiance fields to the data.

\section{\ac{mla} Calibration File}
\label{sec:SM_MLACalibrationFile}

The sequences acquired by the cameras are accompanied by an \ac{mla} calibration file in \texttt{.xml} format. This file is exported from the RxLive~\cite{RxLive} software provided by Raytrix GmbH. However, it has been slightly modified to correspond to the previous standard to allow for better compatibility with other software.

The \ac{mla} calibration file contains information about the position of the \ac{mla}, its orientation, and the characteristics of the micro lenses that make up the \ac{mla}. Raytrix cameras are characterized by having three different types of micro lenses (noted 1, 2, and 3) distributed evenly to form the \ac{mla}. The micro lens data is then separated according to these three types.

\begin{figure}[htbp]
  \centering
  \begin{subfigure}{0.99\linewidth}
    \includegraphics[width=1.0\linewidth]{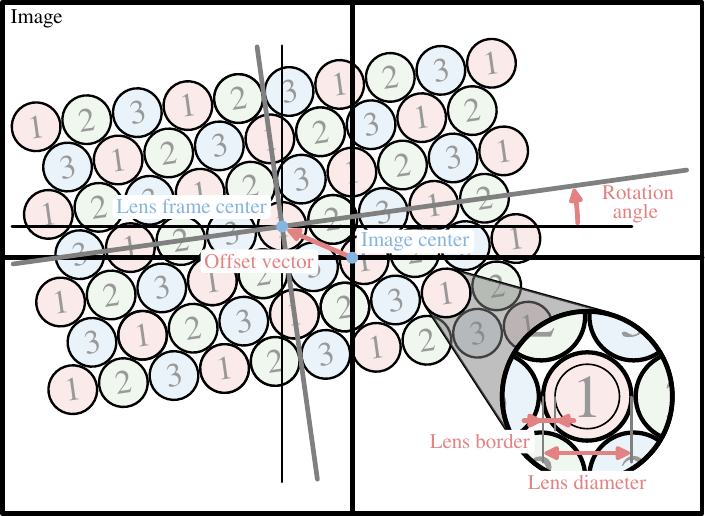}
    \caption{Micro lens parameters and \ac{mla} position and orientation.}
    \label{fig:parametersCalibration-a}
  \end{subfigure}
  \vfill
  \begin{subfigure}{0.52\linewidth}
    \includegraphics[width=1.0\linewidth]{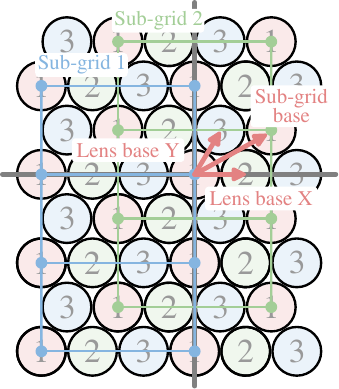}
    \caption{Parameters related to the sub-grids.}
    \label{fig:parametersCalibration-b}
  \end{subfigure}
  \hfill
  \begin{subfigure}{0.46\linewidth}
    \includegraphics[width=1.0\linewidth]{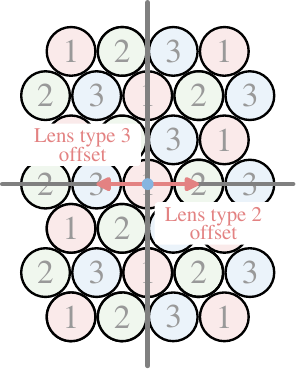}
    \caption{Lens type offset parameters.}
    \label{fig:parametersCalibration-c}
  \end{subfigure}
  \caption{Graphical representation of the parameters in the \ac{mla} calibration file.}
  \label{fig:parametersCalibration}
\end{figure}

The important sections of the file are as follows, and the parameters are represented graphically in Fig.~\ref{fig:parametersCalibration}:
\begin{itemize}
    \item \textbf{offset}: The translation vector between the center of the image and the coordinate reference frame of the \ac{mla}, which is located at its center in the middle of a type 1 micro lens. The vector is characterized by the distances in the $x$ and $y$ directions in pixels, given that the reference frame has its $x$ axis to the right and its $y$ axis to the top (see Fig.~\ref{fig:parametersCalibration-a}).
    \item \textbf{diameter}: The diameter of the micro images, given in pixels (see Fig.~\ref{fig:parametersCalibration-a}).
    \item \textbf{rotation}: The angle between the reference frame on the image and the reference frame on the \ac{mla} (see Fig.~\ref{fig:parametersCalibration-a}). It is generally very small and close to 0. In the file, it is given in radians.
    \item \textbf{lens\_border}: Indicates the outer part of a micro image that should not be considered (see Fig.~\ref{fig:parametersCalibration-a}). It can be enlarged in such a way as to limit distortions or inaccuracies that may appear at the edge of the micro image.
    \item \textbf{tcp}: The total covering plane defined in~\cite{perwass2012single}. It is the furthest distance from the camera for which a depth measurement can be estimated. It is given in virtual depth units introduced in~\cite{perwass2012single}.
    \item \textbf{lens\_base\_x}: The vector between the type 1 micro lens in the center and the nearest type 2 micro lens in the first quadrant (see Fig.~\ref{fig:parametersCalibration-b}). The unit is given in micro lens diameter.
    \item \textbf{lens\_base\_y}: The vector between the type 1 micro lens in the center and the nearest type 3 micro lens in the first quadrant (see Fig.~\ref{fig:parametersCalibration-b}). The unit is given in micro lens diameter.
    \item \textbf{sub\_grid\_base}: The lenses of each type are aligned in a hexagonal grid. It can be divided into two orthogonal grids, called sub-grids. It describes the vector between a micro lens and the closest micro lens of the same type in the first quadrant located on the other rectangular grid (see Fig.~\ref{fig:parametersCalibration-b}). The unit is given in micro lens diameter.
    \item \textbf{lens\_type}: For each of the three types of micro lenses, the \textbf{offset} and \textbf{depth\_range} parameters are given. The \textbf{offset} parameter corresponds to the relative position of the micro lens type with respect to type 1 in lens diameter units (see Fig.~\ref{fig:parametersCalibration-c}). The \textbf{depth\_range} parameter specifies the depth range that can be measured with the lens type in question, expressed in virtual depth units.
\end{itemize}

\section{Data Alignment for the evaluation}
\label{sec:SM_DataAlignment}

This section elaborates on the alignment of different spatial pose data into a unified reference system. The aim is both to enable evaluation by comparing measurements with ground truth and to provide a more intuitive visualization.

First, all data were homogenized to be represented in the same type of coordinate system (a right-handed coordinate system) and with the same transformation representation (transformation matrices in $SE(3)$ where used). It should be noted that Vicon data is recorded in a left-handed coordinate system and must therefore be transformed accordingly.

\begin{figure}[t]
  \centering
  \fbox{\includegraphics[width=0.95\linewidth]{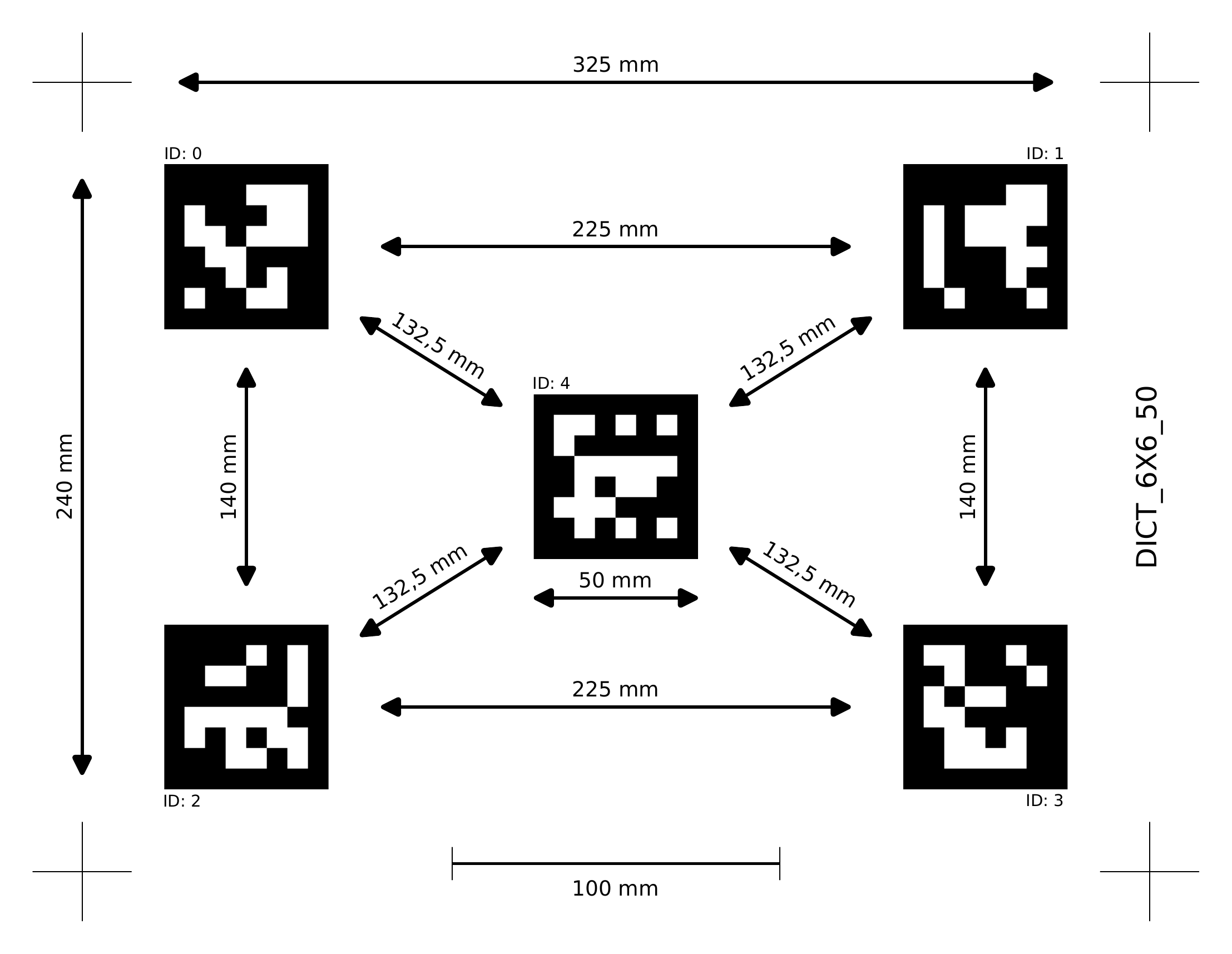}}

   \caption{Pattern of the marker plate used in the scenes of the dataset.}
   \label{fig:patternPlate}
\end{figure}

For the transformations, the lower-left corner of the marker plate was chosen as the reference point. The ArUco markers were used to compute a reference plane. The translation between the ArUco markers and the Vicon markers is known from the marker template shown in Fig.~\ref{fig:patternPlate}. To prevent warping, the template was printed on a rigid foam board. The final transformation into the Vicon reference system is then based on the ArUco marker detections.

The center point of the ArUco marker with ID $X$ (see Fig.~\ref{fig:patternPlate}) is denoted $P_X'=[x_X', y_X', z_X']^T$. To define a local 3D coordinate frame from the ArUco markers, the four markers present on the board are detected: $P_0$, $P_1$, $P_2$, and $P_3$. Marker $P_2$ is chosen as the origin, while $P_2$ and $P_0$ define the directions of the local axes, and $P_1$ serves as a validation point.

The axes of the plane are computed as follows:
\begin{align}
\mathbf{x} &= \frac{P_3 - P_2}{\|P_3 - P_2\|}, \\
\mathbf{y} &= \frac{P_0 - P_2}{\|P_0 - P_2\|}, \\
\mathbf{z} &= \frac{\mathbf{x} \times \mathbf{y}}{\|\mathbf{x} \times \mathbf{y}\|}, \\
\mathbf{z} &\gets \mathbf{x} \times \mathbf{y} \quad \text{(re-orthogonalization)}.
\end{align}

We define the transformation $(R, T)$ between the data from one camera and the common coordinate frame, where $R$ is the rotation matrix and $T$ the translation vector. This enables all data to be represented in the same frame. The axes of the plane form the rotation matrix $R$ according to Eq.~\ref{eq:R} and the translation $T$ corresponds to the position of the origin marker $P_2$, computed with Eq.~\ref{eq:T}.

\begin{equation}
R = [\mathbf{x} \; \mathbf{y} \; \mathbf{z}].
\label{eq:R}
\end{equation}

\begin{equation}
T = P_2.
\label{eq:T}
\end{equation}

To validate the plane, the position of marker $P_1$ is transformed into the local frame using Eq.~\ref{eq:P1_local} and compared with the expected coordinates. The resulting transformation $(R, T)$ defines a stable, right-handed coordinate frame aligned with the marker plane.

\begin{equation}
P_1^{\text{local}} = R^\top (P_1 - P_2)
\label{eq:P1_local}
\end{equation}

Finally, the translation between the lower-left Vicon marker and the lower-left ArUco marker is established. Using this transformation, all visual data can now be aligned with the Vicon ground truth.

\section{Evaluation of the difference between both benchmark methods}
\label{sec:SM_EvaluationdDifference}

Two benchmark methods were evaluated on our dataset: a registration method based on a 3D RANSAC algorithm and a method based on a plenoptic PnP algorithm. Both were assessed against ground truth in the main paper. In this section, we evaluate the difference in pose estimation between the two methods.

Table~\ref{tab:resultsDifferences} highlights the differences in pose estimation in translation and rotation estimates between the 3D RANSAC and plenoptic PnP methods. The 3D RANSAC method benefits from a larger amount of input data. In fact, it takes advantage of a complete point cloud from the calibration of the camera to be registered. On the other hand, the PnP method uses only a single image as input, which considerably reduces the available information but also the complexity of data acquisition. Despite this reduction in input data, the PnP method remains very close to the estimation obtained by 3D RANSAC with an overall \ac{rmse} of 13.61~mm and 0.74°.


\begin{table}[t]
\caption{Translation and rotation differences in the pose estimation of the two benchmark methods (3D RANSAC and plenoptic PnP).}
\label{tab:resultsDifferences}
\centering
\begin{tabular}{|L{2.3cm}|*{8}{C{1.5cm}|}}
\hline
\multirow{2}{*}{\parbox{2.3cm}{\centering Sequence}} & \multicolumn{3}{c|}{Translation difference [mm]} & \multicolumn{3}{c|}{Rotation difference [°]} \\
& RMSE & Mean & SD & RMSE & Mean & SD \\
\hline
00\_Plants & 12.33 & 10.44 & 6.73 & 0.54 & 0.46 & 0.29 \\
01\_Bike & 14.06 & 12.84 & 5.86 & 0.64 & 0.59 & 0.23 \\
02\_Office & 22.93 & 21.11 & 9.21 & 1.49 & 1.35 & 0.65 \\
03\_Electronics & 16.81 & 14.49 & 8.76 & 0.86 & 0.75 & 0.43 \\
04\_Oscilloscope & 4.37 & 4.12 & 1.48 & 0.33 & 0.29 & 0.16 \\
05\_Skeleton & 11.46 & 10.53 & 4.62 & 0.46 & 0.45 & 0.10 \\
06\_Tools & 10.11 & 9.07 & 4.59 & 0.53 & 0.50 & 0.17 \\
\hline
Overall & 13.61 & 11.21 & 7.75 & 0.74 & 0.59 & 0.44 \\
\hline
\end{tabular}
\end{table}

\section{Origin of a systematic offset in absolute translation errors}
\label{sec:SM_systematicOffset}

The reference frame for the Vicon system is located at the center of gravity of the cluster of the four infrared reflective spheres used for camera identification. However, the two proposed methods use the optical center of the camera as a reference. Therefore, a discrepancy exists between the reference frame used by the Vicon tracking system and the frame used for camera recordings.
Table~\ref{tab:resultsEvaluationdDifference} shows the absolute errors separated along the $x$, $y$, and $z$ axes as an example for sequence 00\_Plants in the dataset (the behavior is identical for the other sequences). The $z$ axis corresponds to the optical axis, the $x$ axis is horizontal, and the $y$ axis is vertical relative to the camera. The rotation errors are very small: below 2.32° for the 3D RANSAC method and below 2.13° for the PnP method along each axis. The translation errors along the different axes allow to draw the following conclusions:
\begin{itemize}
    \item The error of 23.01~mm in the direction of the $x$ axis highlights that the coordinate markers are very close along this axis, so the symmetric plane of the camera (depending on the positions of the spheres, the center of gravity may not be exactly centered).
    \item The error of 88.45~mm along the $y$ axis corresponds to the vertical distance between the cluster of spheres and the optical axis of the camera.
    \item The error of 123.69~mm along the $z$ axis indicates the position of the optical center of the camera along the optical axis.
\end{itemize}

Thus, the translation errors can clearly be interpreted as the offset between the coordinate reference used by Vicon and the optical center of the camera used by both registration methods. Through non-linear optimization, this error could be corrected in order to eliminate the systematic offset between the two references and obtain more accurate results.

\begin{table}[t]
\caption{Absolute translation and rotation errors of the pose estimation in the 00\_Plants sequence from the dataset, split along the $x$, $y$, and $z$ axes.}
\label{tab:resultsEvaluationdDifference}
\centering
\begin{tabular}{|C{1.7cm}|C{1.9cm}|C{1.9cm}|C{1.9cm}|C{1.9cm}|}
\hline
\multirow{2}{*}{\parbox{1.9cm}{\centering Metric}} & \multicolumn{2}{c|}{3D RANSAC Method} & \multicolumn{2}{c|}{Plenoptic PnP Method} \\
& \textbf{RMSE} & \textbf{SD} & \textbf{RMSE} & \textbf{SD} \\
\hline
\multicolumn{5}{|c|}{Translation absolute error} \\
\hline
$x$ [mm]  & 23.01 & 17.54 & 25.36 & 27.77 \\
$y$ [mm]  & 88.45 & 31.32 & 89.98 & 37.08 \\
$z$ [mm]  & 123.69 & 7.59 & 127.90 & 6.98 \\
\hline
\multicolumn{5}{|c|}{Rotation absolute error} \\
\hline
$x$ [°] & 1.72 & 1.07 & 2.13 & 0.48 \\
$y$ [°] & 2.32 & 0.99 & 1.52 & 0.66 \\
$z$ [°] & 2.07 & 1.03 & 2.01 & 1.04 \\
\hline
\end{tabular}
\end{table}

\end{document}